\documentclass{sig-alternate}

\newfont{\mycrnotice}{ptmr8t at 7pt}
\newfont{\myconfname}{ptmri8t at 7pt}
\let\crnotice\mycrnotice%
\let\confname\myconfname%

\toappear{}

\usepackage{verbatim}

\usepackage[overlay,absolute]{textpos}

\begin{document}

\begin{textblock*}{10in}(16mm, 10mm)
{\textbf{Ref:} \emph{ACM Genetic and Evolutionary Computation Conference (GECCO)}, pages 1451--1452, Vancouver, Canada, July 2014.}
\end{textblock*}

\clubpenalty=10000
\widowpenalty = 10000

\title{Genetic Algorithms for Evolving Deep Neural Networks}

\numberofauthors{2}

\author{
% 1st. author
\alignauthor
Eli (Omid) David\\
       \affaddr{Bar-Ilan University}\\
       \affaddr{Ramat-Gan 52900, Israel}\\
       \email{mail@elidavid.com}\\ 
% 2nd. author
\alignauthor
Iddo Greental\\
       \affaddr{Tel Aviv University}\\ 
       \affaddr{Tel Aviv 69978, Israel}\\
       \email{iddo.greental@gmail.com}\\
}

\maketitle

\begin{abstract}
In recent years, \emph{deep learning} methods applying unsupervised learning to train deep layers of neural networks have achieved remarkable results in numerous fields. In the past, many genetic algorithms based methods have been successfully applied to training neural networks. In this paper, we extend previous work and propose a GA-assisted method for deep learning. Our experimental results indicate that this GA-assisted approach improves the performance of a deep autoencoder, producing a sparser neural network.
\end{abstract}

%\category{I.2.6}{Artificial Intelligence}{Learning}[Connectionism and neural nets]

%\terms{Algorithms}

%\keywords{Genetic algorithms, Deep learning, Neural networks, Autoencoders}

\vspace{1mm}
\noindent
{\bf Categories and Subject Descriptors:} I.2.6 {[Artificial Intelligence]}: {Learning}---\emph{Connectionism and neural nets}

\vspace{1mm}
\noindent
{\bf General Terms:} Algorithms.

\vspace{1mm}

\section{Introduction}

While the motivation for creating deep neural networks consisting of several hidden layers has been present for many years, supported by a growing body of knowledge on the deep architecture of the brain and advocated on solid theoretical grounds \cite{bengio07a,bengio09}, until recently it was very difficult to train neural networks with more than one or two hidden layers.

Recently, \emph{deep learning} methods which facilitate the training of neural networks with several hidden layers have been the subject of increased interest, owing to the discovery of several novel methods. Common approaches employ either \emph{autoencoders} \cite{bengio07b,ranzato07} or \emph{restricted Boltzmann machines} \cite{hinton06a,hinton06b,lee08} to train one layer at a time in an unsupervised manner.

In the past, genetic algorithms have been applied successfully to training neural networks of shallow depths (one or two hidden layers) \cite{schaffer92}. In this paper we demonstrate how genetic algorithms can be applied to improve the training of deep autoencoders.

\section{Deep Autoencoders}

In this section, we briefly describe autoencoders and explain how they are used in the context of deep learning.

An autoencoder is an unsupervised neural network which sets the target values (of the output layer) to be equal to the inputs, i.e., the number of neurons at the input and output layers is equal, and the optimization goal for output neuron $i$ is set to $y_i = x_i$, where $x_i$ is the value of the input neuron $i$. A hidden layer of neurons is used between the input and output layers, and the number of neurons in the hidden layer is usually set to fewer than those in the input and output layers, thus creating a bottleneck, with the intention of forcing the network to learn a higher level representation of the input. The weights of the encoder layer ($W$) and the weights of the decoder layer ($W'$) can be \emph{tied} (i.e., defining $W' = W^T$).

Autoencoders are typically trained using backpropagation. When an autoencoder's training is completed, we can discard the decoder layer, fix the values of the encoder layer (so the layer can no longer be modified), and treat the outputs of the hidden layer as the inputs to a new autoencoder added on top of the previous autoencoder. This new autoencoder can be trained similarly. Using such layer-wise unsupervised training, deep stacks of autoencoders can be assembled to create deep neural networks consisting of several hidden layers (forming a \emph{deep belief network}). Given an input, it will be passed through this deep network, resulting in high level outputs. In a typical implementation, the outputs may then be used for supervised classification if required, serving as a compact higher level representation of the data.

\section{GA-Assisted Deep Learning}

Genetic algorithms (GA) have been successfully employed for training neural networks \cite{schaffer92}. Specifically, GAs have been used as substitute for the backpropagation algorithm, or used in conjunction with backpropagation to improve the overall performance.

We now propose a simple GA-assisted method which (according to our initial results presented in the next section) improves the performance of an autoencoder, and produces a sparser network.

When training an autoencoder with tied weights (i.e., the weights of the encoding layer are tied to those of the decoding layer), we store multiple sets of weights ($W$) for the layer. That is, in our GA population each chromosome is one set of weights for the autoencoder. For each chromosome (which represents the weights of an autoencoder), the root mean squared error (RMSE) is calculated for the training samples (the error for each training sample is defined as the difference between the values of the input and output layers). The fitness for chromosome $i$ is defined as $f_i = 1 / RMSE_i$. After calculating the fitness score for all the chromosomes, they are sorted from the fittest to the least fit. The weights of the high ranking chromosomes are updated using backpropagation, and the lower ranking chromosomes are removed from the population. The removed chromosomes are replaced by the offsprings of the high ranking chromosomes. The selection is performed uniformly with each of the remaining chromosomes having an equal probability for selection (regardless of the fitness values of the chromosomes, i.e., the fitness score is used only for determining which chromosomes are removed from the population). Given two parents, one offspring is created as follows: Crossover is performed by randomly selecting weights from the parents, and mutation is performed by replacing a small number of weights with zero.

Gradient descent methods such as backpropagation are susceptible to trapping at local minima. Our method assists backpropagation in this respect, reducing the probability of trapping at local minima. Additionally, mutating the weights to zero encourages sparsity in the network (fewer active weights). Sparse representations are appealing due to information disentangling, efficient variable-size representation, linear separability, and distributed sparsity \cite{glorot11}. 

Note that when training of an autoencoder is complete, the values of the best chromosome are selected for that autoencoder. These values are fixed and shared amongst all chromosomes when a new autoencoder layer is added on top of the previously trained layer. Thus, each chromosome contains only the values of the layer currently being trained.

\section{Experimental Results}

For our experiments we used the popular MNIST handwritten digit recognition database \cite{lecun98}. In the MNIST dataset, each sample contains 784 pixels (28x28 image), each having a grayscale value between 0 to 255 (which we scale to a 0 to 1 range). Each sample also contains a target classification label (between 0 and 9), which is used for the subsequent supervised classification phase (using the high level representations generated by the unsupervised autoencoder).

Our deep neural network uses a stack of 5 layers. The first layer has 784 neurons, followed by four higher level layers consisting of 500, 250, 100, and 50 neurons. Each layer is trained separately, with the next layer added only once training is complete: first we train the 784-500 layer, then use the 500 output neurons as inputs to the 500-250 layer, and similarly for the 250-100 and 100-50 layers.

The GA implementation uses a population of 10 chromosomes. In each generation, the five worst chromosomes (half the population) are removed and replaced by the offsprings of the five best chromosomes. We used crossover and mutation rates of 0.8 and 0.01 accordingly.

To compare the performance of our GA-assisted method with traditional backpropagation, we ran both methods under similar conditions. First, we ran the traditional backpropagation version 10 times and selected the result with the least reconstruction error (best tuned). Next, we ran the GA-assisted method only once, allowing the same total runtime as the previous method. Comparing the reconstruction errors of the two approaches, the GA-assisted method consistently yielded a smaller reconstruction error, as well as a sparser network.

In order to compare the classification accuracy of the two methods, we ran 10,000 new test samples through the two trained networks and recorded the 50 output values for each sample. Recall that in this test phase the weights of the network are already fixed, hence an input sample of 784 values is passed through the layers of 500, 250, 100, and 50 neurons without modifying their weights. The representation quality of the networks can be compared by applying supervised classification to the higher level values
produced by the 50 neurons of the output layer. We used SVM classification with a radial basis function (RBF) kernel. Using SVM, the traditional autoencoder achieved a 1.85\% classification error, while the GA-assisted method's classification error was 1.44\%.

\section{Concluding Remarks}

In this paper we presented a simple GA-assisted approach, which according to our initial results improves the performance of a deep autoencoder. While our implementation used an autoencoder, the same method is applicable to other forms of deep learning such as restricted Boltzmann machines (RBM).

In recent years, several improvements upon traditional autoencoders and RBM have been proposed which improve their generalization. Such improvements include \emph{dropout} \cite{hinton12}, which randomly disables some neurons during training, \emph{dropconnect} \cite{wan13}, which randomly disables some weights during training, and \emph{denoising autoencoders} \cite{vincent10}, which randomly add noise by removing a portion of the training data. The improved performance of the GA-assisted autoencoder could arise from a similar principle, since mutation randomly disables some of the weights during training. It is important to compare the GA-assisted approach to the above mentioned alternative improvements in future research.

\end{document}